\documentclass[11pt]{article}

\usepackage{acl}

\usepackage{times}
\usepackage{latexsym}

\usepackage{amsmath} 

\usepackage[T1]{fontenc}
\usepackage[utf8]{inputenc}

\usepackage{microtype}

\usepackage{inconsolata}

\usepackage{graphicx}

\usepackage{float}

\graphicspath{{images/}}

\title{Evaluating Small Decoder-Only Language Models for Grammar Correction and Text Simplification}

\author{
  Anthony Lamelas $\bullet$ 
  New York University\\
  \texttt{al8372@nyu.edu}
}

\begin{document}
\maketitle
\begin{abstract}
Large language models have become extremely popular recently due to their ability to achieve strong performance on a variety of tasks, such as text generation and rewriting, but their size and computation cost make them difficult to access, deploy, and secure in many settings. This paper investigates whether small, decoder-only language models can provide an efficient alternative for the tasks of grammar correction and text simplification. The experiments in this paper focus on testing small language models out of the box, fine-tuned, and run sequentially on the JFLEG and ASSET datasets using established metrics. The results show that while SLMs may learn certain behaviors well, their performance remains below strong baselines and current LLMs. The results also show that SLMs struggle with retaining meaning and hallucinations. These findings suggest that despite their efficiency advantages, current SLMs are not yet competitive enough with modern LLMs for rewriting, and further advances in training are required for SLMs to close the performance gap between them and today's LLMs.
\end{abstract}
\section{Introduction}

Grammar correction and text simplification are very important natural language processing (NLP) tasks, specifically for writing assistance, accessibility, and text analysis. Many high-quality systems today rely on extremely large language models (LLMs). Because of this, they are very expensive to train, require substantial computing power, cannot be hosted locally, and are costly in terms of energy and memory. 

On the other hand, small language models (SLMs) provide a potential alternative as they are often open source (making them accessible, able to run locally, and easy to fine-tune) and cost-efficient due to their size. They also require much less computing power to both train and fine-tune. However, their performance on tasks such as grammar correction and text simplification is underexplored. 

Despite interest in SLMs, there has been little to no systematic evaluation of fine-tuned SLMs specifically for grammar correction and text simplification. It is also unclear how well task-specific fine-tuning improves performance, leaving the question of whether SLMs can compete with LLMs up in the air. 

Another open question is whether applying multiple SLMs sequentially (cascading), where each model improves upon the results of the previous, can improve performance beyond that of a single SLM. In this work, we fine-tuned several SLMs on both grammar correction and text simplification, then evaluated them using standard and widely accepted benchmarks. We then compared these SLMs (singular and cascading) directly to an LLM.

\section{Related Work}

\subsection{Scaling, Model Size, and Efficiency}
Past language modeling research has typically shown performance gain through increasing model size. Quality has a direct correlation with an increase in data, compute resources, and parameters  \cite{kaplan2020scaling}. However, there are some examples of more recent work that have highlighted the limitations of that strategy. For example, some researchers have focused on performance per unit of compute \cite{hoffmann2022training}. These developments have motivated interest in smaller model architectures that are easier to fine-tune, less expensive, and more practical for specific applications.

\subsection{Encoder--Decoder and Decoder-Only Approaches}
Text rewriting tasks have typically been modeled with encoder-decoder architectures in the past. Models such as BART \cite{lewis2019bart} and T5 \cite{raffel2020exploring} are examples of such models that have performed edits well across tasks such as translation, summarization, and correction.  Decoder-only models were initially considered less suitable for complex editing \cite{rothe2020leveraging}; however, some recent work has shown that some of these models were able to write quite well, such as TinyLlama 1.1B \cite{eldan2023tinyllama}. Because of this, interest in evaluating decoder-only models has increased, even though encoder-decoder systems have dominated most past research. 

\subsection{Grammar Error Correction}
Systematic grammar error correction (GEC) began with rule-based systems, then moved to statistical models, and currently, mostly neural networks are used. Early GEC systems relied on encoder--decoder architectures trained on large datasets. These models provided a foundation for training data and the establishment of benchmarks such as M2 and Generalized Language Evaluation Understanding (GLEU). While encoder--decoder models, such as BART and T5, have achieved strong performance \cite{lewis2019bart, raffel2020exploring}, small decoder-only models have remained mostly unexplored for GEC, despite recent evidence that smaller causal models can acquire strong rewriting abilities \cite{eldan2023tinyllama}.

\subsection{Text Simplification Research}
Many encoder--decoder architectures have been trained on datasets such as WikiAuto \cite{jiang2020wikiauto} and ASSET \cite{alva2020asset}. ASSET provides multiple human references per sentence and established SARI as a standard metric for evaluating structural and content-preserving simplification. Systems such as ACCESS \cite{martin2020access} and MUSS \cite{martin2021muss} are examples that apply pre-trained encoder--decoder models to get strong results. However, similar to GEC, small decoder-only models have received almost no attention regarding modern simplification benchmarks, leaving an open question.

\subsection{Iterative Refinement and Cascading}
Iterative refinement is another direction for rewriting tasks. Methods focusing on refining a model's own output show that models can improve their output quality by repeatedly giving feedback and then revising \cite{madaan2023self}. Practices like this have been revolutionary in the world of LLMs, increasing the quality of output by allowing them to "think" about their responses. It is quite common to see thinking modes from many of the top LLM providers. This refinement is typically all done using a single large model, but far less is known about whether it will work if distributed across several small models.

\subsection{Research Gap}
Although encoder--decoder models, LLMs, and iterative refinement techniques have been studied extensively,  few works have studied how fine-tuned small decoder-only SLMs perform on grammar correction and simplification benchmarks, or whether multi-stage refinement with these models provides any benefit.

\section{Data}

For the grammar correction task, the BEA-2019 dataset was used for both training (68,816 examples) and validation (8,768 examples). BEA-2019 contains a large set of sentences annotated with grammatical error corrections \cite{bryant2019bea}. The corrections include many diverse examples such as verb form, prepositions, agreement, spelling, and more \cite{bryant2019bea}. This dataset is also quite popular in the NLP research world and is one of the standard datasets for supervised training of grammar correction models. 

For testing the grammar correction, the JFLEG dataset was used (748 examples). This dataset includes rewrites of sentences that focus on fluency and correctness \cite{napoles2017jfleg}. This is another widely used dataset as it measures both correctness and the naturalness of writing, making it more challenging than purely error-tagged datasets \cite{napoles2017jfleg}.

For the text simplification task, the WikiAuto dataset was used for training (373,801 examples). This dataset provides pairs of sentences extracted from Wikipedia \cite{jiang2020wikiauto}. One sentence within these pairs is considered "normal", and the other is the simplified version \cite{jiang2020wikiauto}. When the model is given the normal pair, its job is to simplify it, making it very easy to compare the results to the simplified version \cite{jiang2020wikiauto}.

For validation (2000 examples) and testing (359 examples) of the text simplification, the ASSET dataset was used. This dataset has multiple human-written simplifications per sentence \cite{alva2020asset}. ASSET was used as it captures paraphrasing, compression, and meaning preservation \cite{alva2020asset}. These qualities make ASSET versatile during the validation phase.

\section{Methodology}

\subsection{Task}

As mentioned earlier, two tasks were studied these being grammar correction and text simplification. It's important to note that these tasks use separate training, validation, and test datasets, as well as separate fine-tuned models.

\subsection{Models}

Five different SLMs were tested, and three different SLMS were fine-tuned. GPT-2, GPT-2 Medium, and GPT-Neo-125M were fine-tuned. Gemma 2B and TinyLlama 1.1B were not. All of the models were used in causal-language-modeling mode (meaning the models cannot cheat by looking at future words; they only predict the next token based on the previous ones). GPT-4 and GPT-3.5 Turbo were the baseline LLMs used for comparison. The LLM baselines were accessed via the OpenAI API \cite{openai2024models}.

We decided to only test decoder-only SLMs. A few reasons for this are keeping consistency with modern LLMs as they are also decoder-only,  efficiency (do not have to encode, so memory is saved), and to focus only on general language modelling, which is the overarching field of our research. 

\subsection{Training and Setup}

As mentioned previously in the data section, the grammar correction models used the BEA-2019 dataset for training and validation, and the JFLEG dataset for testing. The text simplification models used the WikiAuto dataset for training and the ASSET dataset for validation and testing.

The Hugging Face Transformers (HFT) framework was used for the training setup. All of the models were fine-tuned using a causal-language-modeling setup. We used mostly similar training parameters across both tasks (epochs, optimizer, learning rate, and sequence length were mostly the same across models). Full hyperparameter details are reported in Appendix A.

Models were fine-tuned and evaluated on the Greene cluster on NYU's HPC. The GPUs used were RTX8000s and A100s. Specific hardware allocations can be found in Appendix B.

\subsection{Inference and Decoding}

After fine-tuning, each model generates new text by converting probability distributions into word sequences and choosing the most probable word. The goal of this is to produce text that reflects the original objective. During validation, we observed especially high hallucination rates under higher temperatures. Because of this, we decided to reduce the temperature fully and use a greedy strategy for all final evaluations. The full decoding configurations can be found in Appendix A, and the relevant validation data can be found in Appendix D.

\subsection{Cascading SLM Pipeline}

The reason for the cascading strategy was to determine if applying SLM more than once can improve accuracy. The pipeline functions in two stages. First, the first SLM generates a corrected/simplified output. Then, that output is fed to a second SLM (could be the same model twice) to produce a refined output with the goal of correcting the mistakes made in the first pass. The hyperparameters, decoding, and evaluation remain the same for both the cascading and single-SLM strategies.

\subsection{Evaluation Metrics}

We evaluate our grammar correction and text simplification models using a mix of accuracy-based,
edit-based, fluency-based, and readability-based metrics. Each metric captures a different property
of output quality, and together they provide a comprehensive evaluation across both tasks.

\subsubsection{Grammar Correction Metrics}

\paragraph{GLEU:}
GLEU is designed for grammatical error correction, particularly
for datasets such as JFLEG that contain multiple human rewrites. GLEU averages precision and recall over $n$-grams. We compute this metric using the natural language toolkit (NLTK) implementation.

\paragraph{\textbf{M$^2$ Precision, Recall, and {F$_{0.5}$}}}
For edit-based evaluation, we use M$^2$, which aligns predicted edits to gold edits.  
Precision and recall are defined as:
\begin{align}
    \text{Precision} &= \frac{\text{TP}}{\text{TP} + \text{FP}}, \label{eq:precision} \\
    \text{Recall} &= \frac{\text{TP}}{\text{TP} + \text{FN}}. \label{eq:recall}
\end{align}
The weighted $F_{\beta}$ score is:
\begin{equation}
    F_{\beta} = (1 + \beta^2) \cdot
    \frac{\text{Precision} \cdot \text{Recall}}
         {\beta^2 \cdot \text{Precision} + \text{Recall}}.
\end{equation} 
where ${\beta = 0.5}$.

\subsubsection{Text Simplification Metrics}

\paragraph{SARI}
SARI (\emph{System output Against References and against the Input}) evaluates how well a system 
adds, deletes, and keeps $n$-grams. It is defined as

{
\begin{equation}
\text{SARI} = \frac{1}{3N} \sum_{n=1}^{N} \big( \text{Add}_n + \text{Keep}_n + \text{Delete}_n \big)
\end{equation}
}

Add, delete, and keep are explained as: 
\begin{itemize}
    \item \textbf{Add}: correct new information or simpler phrasing
    \item \textbf{Delete}: removal of unnecessary content
    \item \textbf{Keep}: amount of meaning preserved
\end{itemize}
The final SARI score is the average of these components.

\paragraph{Flesch Reading Ease (FRE)}
This measures text simplicity, defined as:
{\small
\begin{equation}
    \text{FRE} = 206.835 - 1.015 \cdot \frac{\text{words}}{\text{sentences}}
                 - 84.6 \cdot \frac{\text{syllables}}{\text{words}}.
\end{equation}
}
Higher FRE values indicate easier-to-read sentences.

\paragraph{Length and Compression}
We report the average lengths (in words) of the input ($L_{\text{in}}$), prediction ($L_{\text{pred}}$),
and reference sentences ($L_{\text{ref}}$), along with the compression ratio:
\begin{equation}
    \text{Compression} = \frac{L_{\text{pred}}}{L_{\text{in}}}.
\end{equation}
A compression ratio below 1 means that the model shortened the input, while a ratio above 1 indicates lengthening the input. These statistics help contextualize SARI and readability scores, showing whether the model simplifies by compressing or by rewriting without much reduction.

\subsubsection{Hallucination Metric}

\paragraph{Named-Entity Rate (NER)}
To measure whether models add new content that is not originally present in the input, we compute NER using spaCy. Named entities are taken from both the input sentence and the prediction. An entity is considered hallucinated if it appears in the prediction but not the input. Let $H_i$ be an indicator that example $i$ contains at least one hallucinated entity. The hallucination rate is then:
\[
\text{Named-Entity Rate} = \frac{1}{N} \sum_{i=1}^{N} H_i \times 100.
\]

\subsection{Libraries and Tools}

The HFT library was used to load models, tokenize, and finetune the models. HFT also handled mixed precision, gradient checkpointing, and logging. PyTorch was used to handle GPU computation, optimization steps, and bf16/fp16. ERRANT was used to perform automatic grammatical error alignment for the $M^{2}$ metric. NLTK was used for tokenization during preprocessing and to compute the GLEU scores for grammar correction. Textstat was used to compute the FRE and FKGL readability metrics. Pandas was used to load CSV datasets and to save intermediate outputs during cascading. SLURM was used to schedule jobs on the HPC as mentioned above.

\section{Experiments}
\label{sec:bibtex}

Two categories of experiments were run. These were grammar correction and text simplification, both including single-pass and cascaded tests. Each SLM that was tested was fine-tuned separately for each task. GPT-4 and GPT-3.5 Turbo were used as a baseline for comparisons and were tested on the same datasets and evaluated using the same metrics as the SLMs.

All of the fine-tuned models shared the same hyperparameters except for the batch size in a few cases. Cascading was run in two and three stages (2-3 SLMs took a pass on the data). We focused on comparing SLMs vs other SLMs, single-pass vs cascaded strategies, and best SLM output versus the GPT-4 and GPT-3.5 Turbo baselines.

\section{Results}

\subsection{Grammar Correction Results}

\subsubsection{Single-Pass Results}

Figure 1 describes the results of the single-pass grammar correction experiment, where only one model was used at a time. Overall, when evaluating based on the GLEU metric, the LLMs outperformed the SLMs by a large margin.

\begin{figure}[H]
    \centering
    \includegraphics[width=\linewidth]{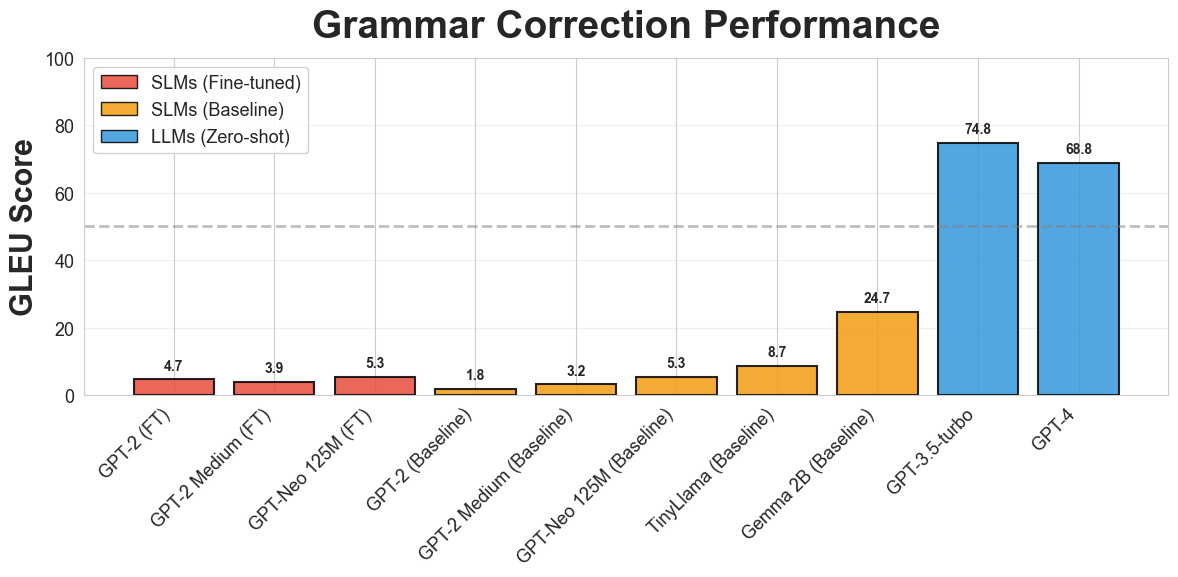}
    \caption{Single-Model GLEU Scores}
    \label{fig:mylabel}
\end{figure}

Figure 2 reports $M^2$  precision, recall, and $F_{0.5}$ scoring. These metrics reflect edit-based performance on the correction of grammar. When evaluated on these metrics, the LLMs again outperformed the SLMs by a large margin.

\begin{figure}[H]
    \centering
    \includegraphics[width=\linewidth]{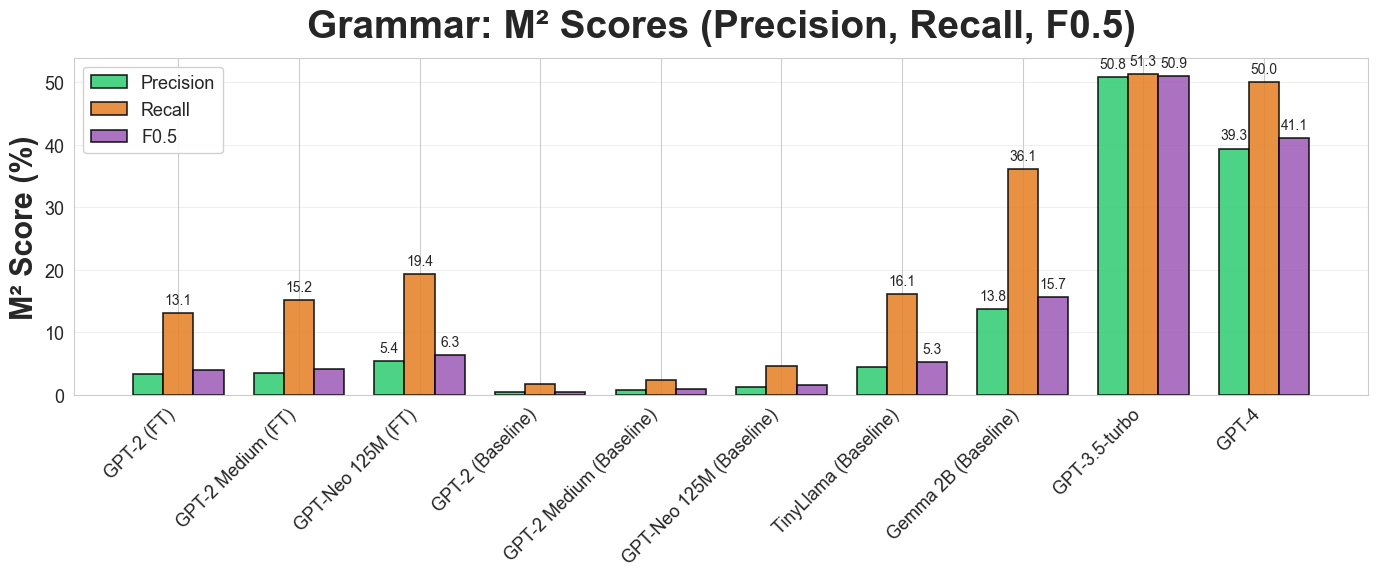}
    \caption{Precision, Recall, and $F_{0.5}$}
    \label{fig:mylabel}
\end{figure}

\subsubsection{Cascading Results}
Figure 3 shows the GLEU scores for all of the cascading jobs that we ran. As you can see, the cascading approach led to no significant changes.

\begin{figure}[H]
    \centering
    \includegraphics[width=\linewidth]{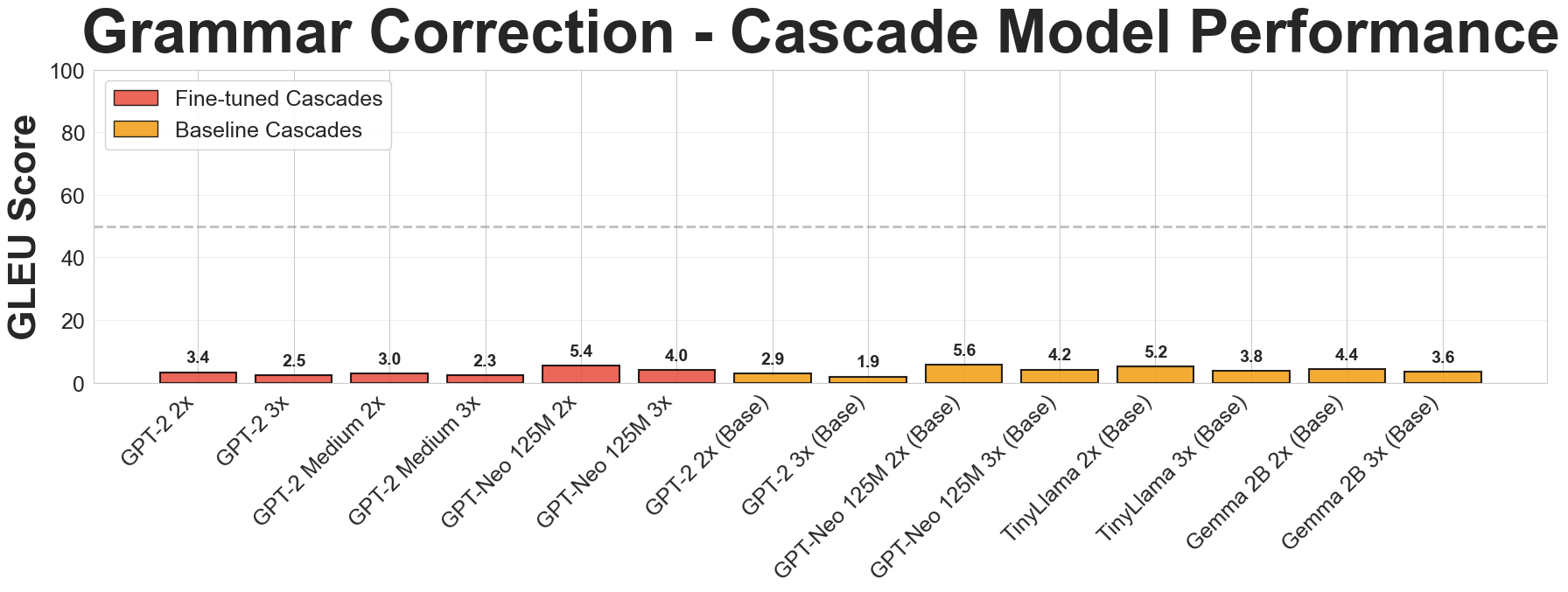}
    \caption{Cascade GLEU Scores}
    \label{fig:mylabel}
\end{figure}

Figure 4 shows the precision, recall, and $F_{0.5}$ scores for all of the cascades. In this case, the fine-tuned models performed much better.

\begin{figure}[H]
    \centering
    \includegraphics[width=\linewidth]{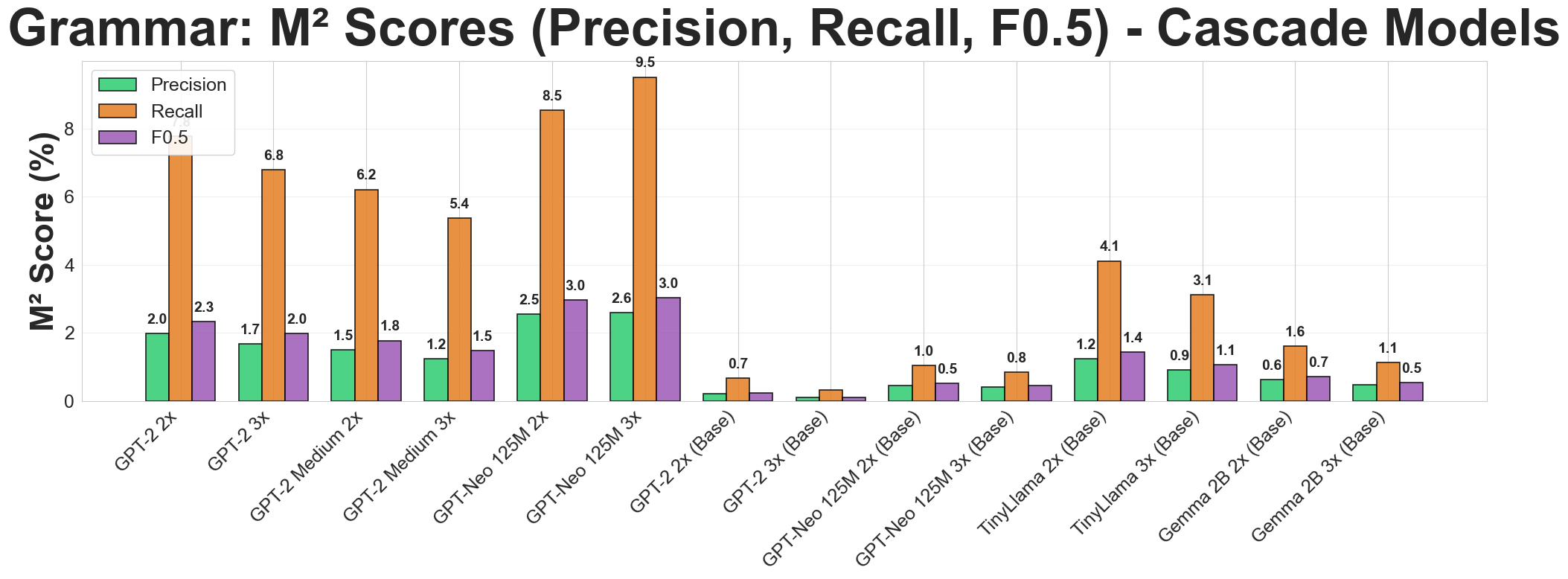}
    \caption{Cascade Precision, Recall, and $F_{0.5}$}
    \label{fig:mylabel}
\end{figure}

\subsection{Text Simplification Results}

\subsubsection{Single-Pass Results}

Figure 5 shows the SARI scores from all of the models when evaluated on the ASSET test dataset. The LLMs achieved much higher scores than the finetuned SLMs but were only slightly better than the baseline SLMs. 

\begin{figure}[H]
    \centering
    \includegraphics[width=\linewidth]{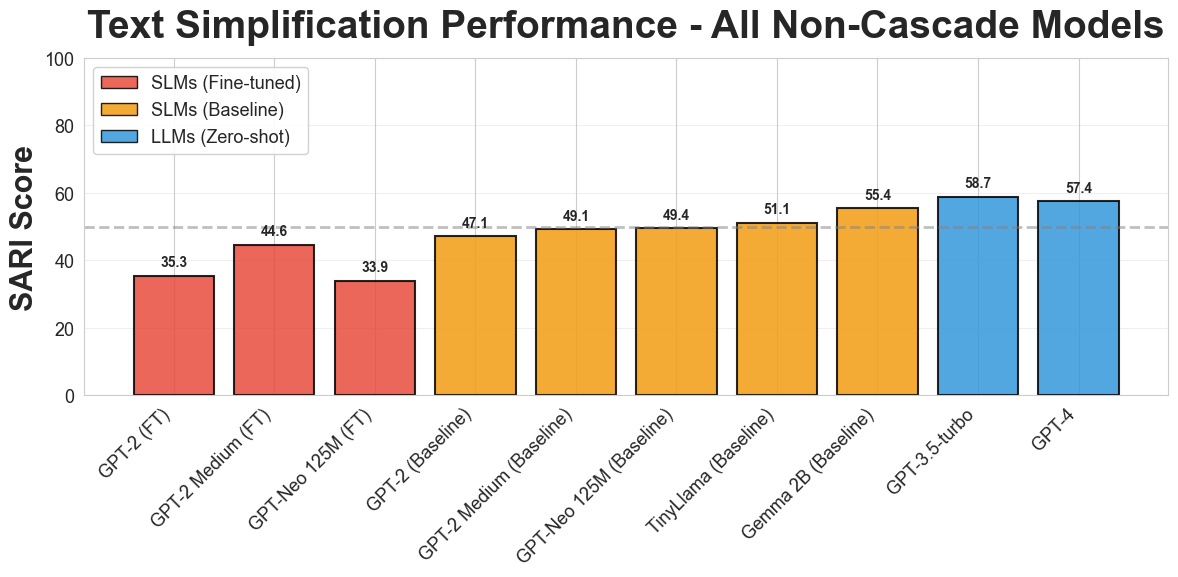}
    \caption{Single-Model SARI Scoring}
    \label{fig:mylabel}
\end{figure}

Figure 6 reports Flesch Reading Ease scores. The LLMs performed well, but both the baseline and fine-tuned SLMs remained competitive.

\begin{figure}[H]
    \centering
    \includegraphics[width=\linewidth]{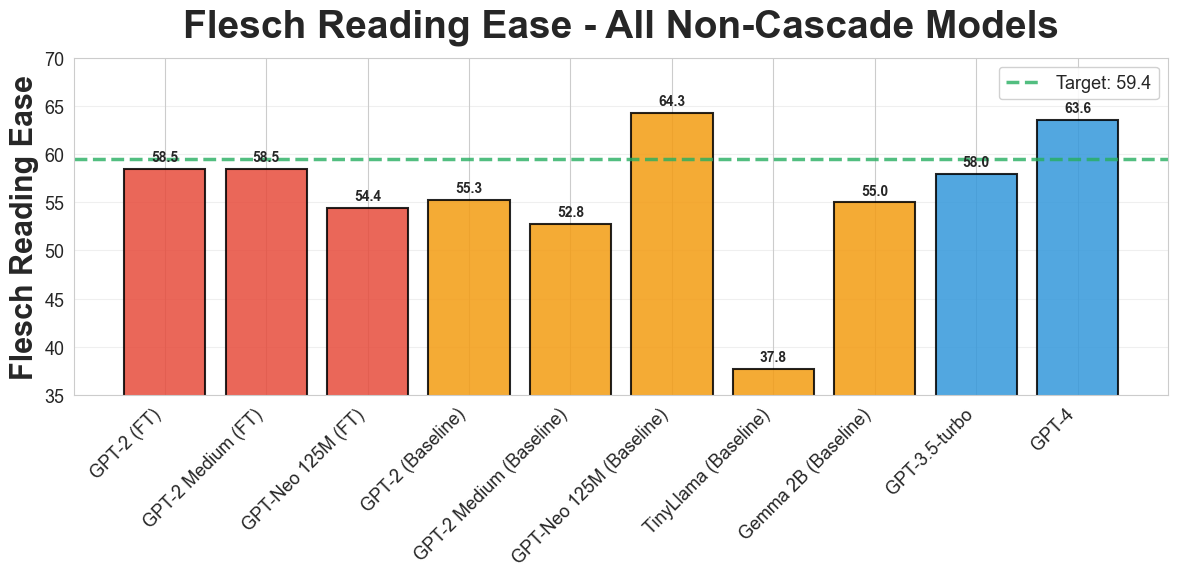}
    \caption{Single-Model Flesch Reading Ease}
    \label{fig:mylabel}
\end{figure}

Figure 7 shows the percentage of compression on the test text. LLMs typically compressed or decompressed the text slightly; however, the SLMs consistently decompressed the text by various percentages.

\begin{figure}[H]
    \centering
    \includegraphics[width=\linewidth]{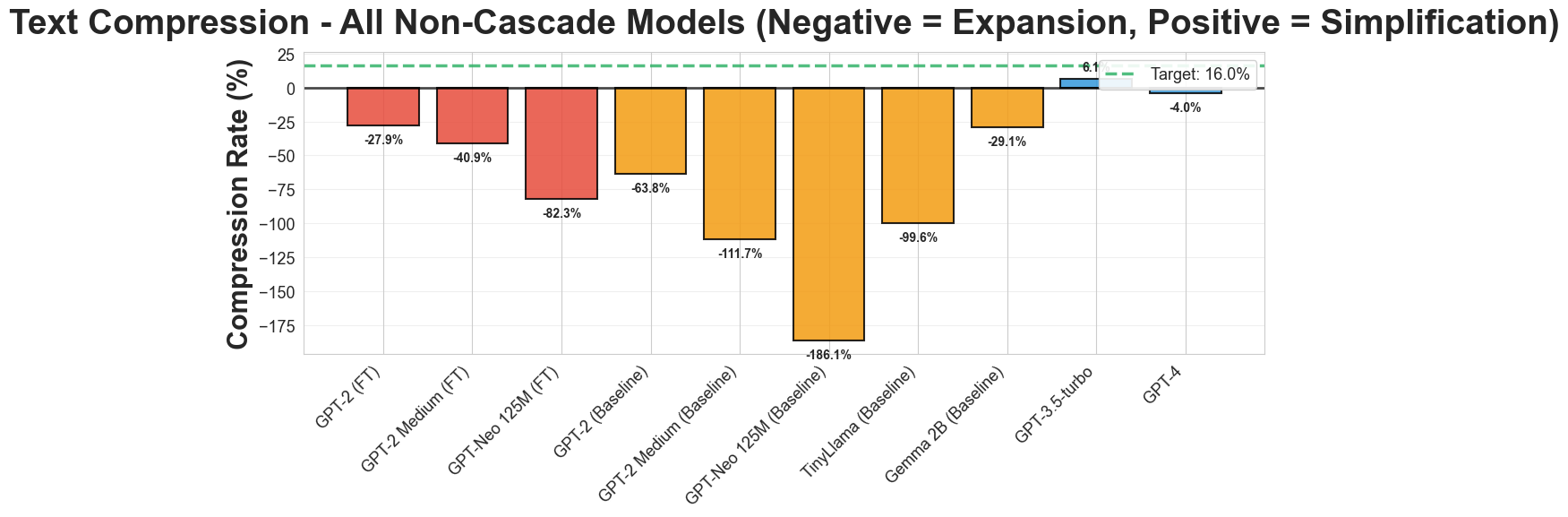}
    \caption{Single-Model Text Compression Rates}
    \label{fig:mylabel}
\end{figure}

\subsubsection{Cascading Results}
Figure 8 shows all of the SARI scores when cascading was performed. As you can see, the baseline cascades performed slightly better than the fine-tuned cascades.

\begin{figure}[H]
    \centering
    \includegraphics[width=\linewidth]{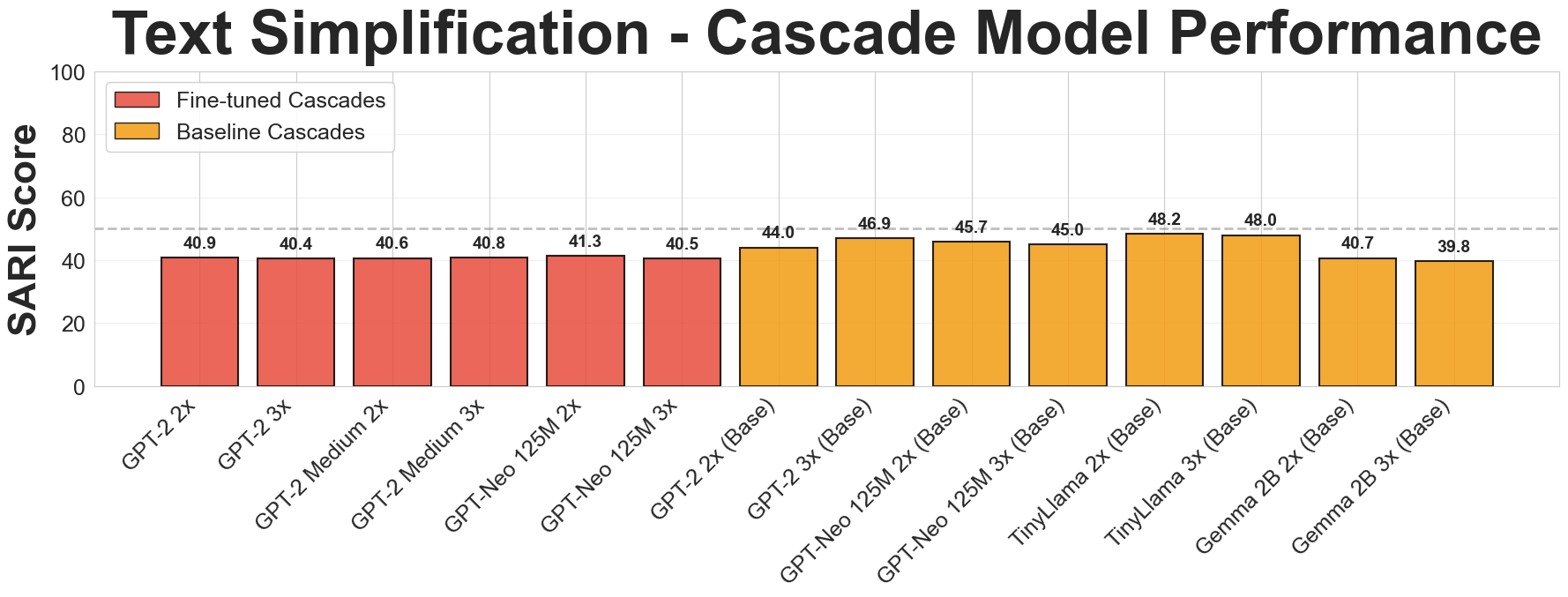}
    \caption{Cascade SARI Scoring}
    \label{fig:mylabel}
\end{figure}

Figure 9 shows the FRE for the cascaded models. These results were inconsistent, having some baseline cascades performing better than the fine-tuned cascades, but others performing worse.

\begin{figure}[H]
    \centering
    \includegraphics[width=\linewidth]{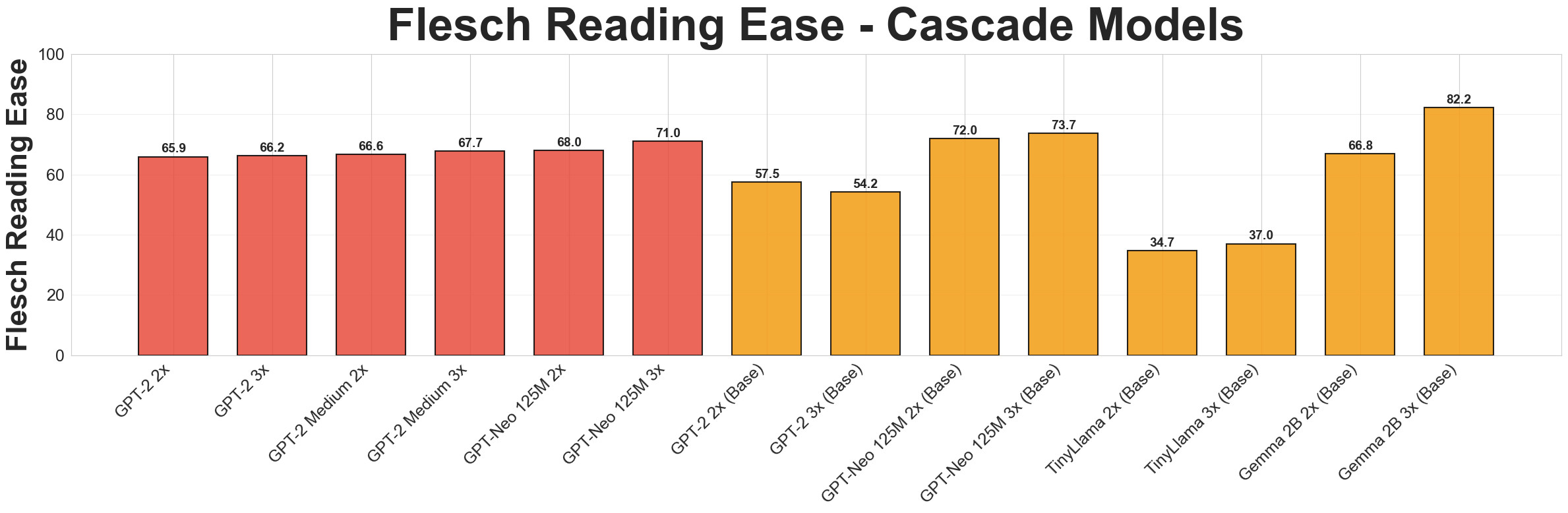}
    \caption{Cascade Flesch Reading Ease}
    \label{fig:mylabel}
\end{figure}

Figure 10 shows the text compression rates for the cascading models. As you can see, the finetuned models did a much better job of compressing text than the inconsistent baseline models.

\begin{figure}[H]
    \centering
    \includegraphics[width=\linewidth]{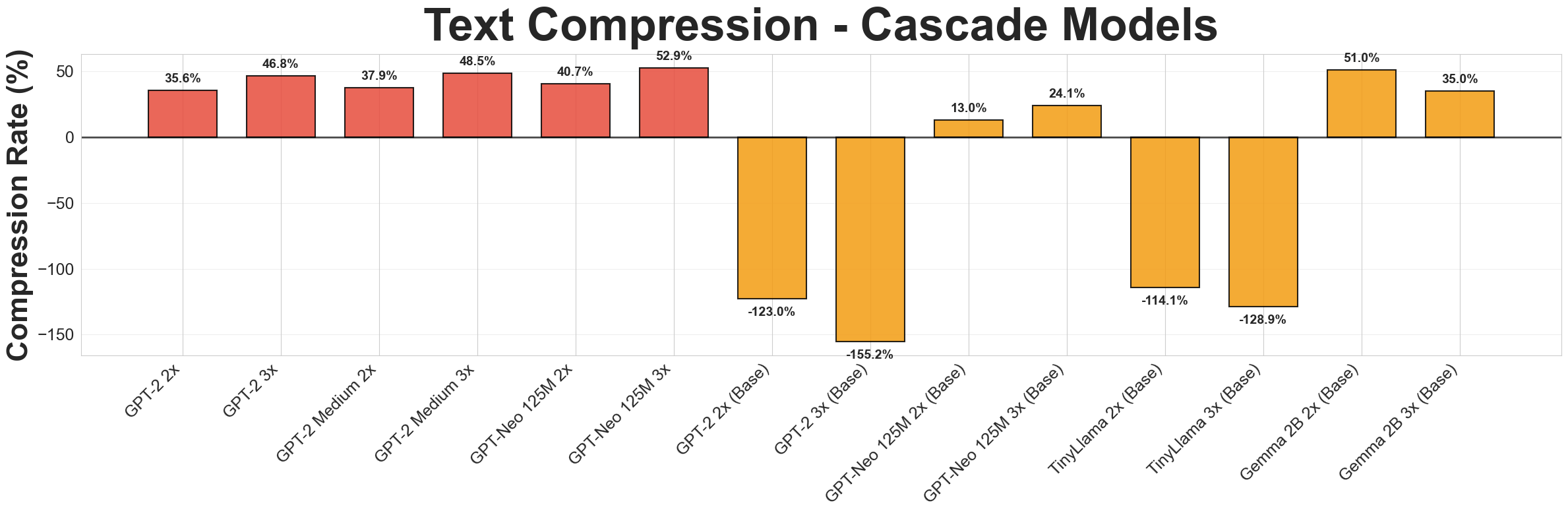}
    \caption{Cascade Text Compression Rates}
    \label{fig:mylabel}
\end{figure}

\subsection{Hallucination Results}
Figure 11 shows how much all of the implementations hallucinated when correcting grammar. As you can see, the LLMs and baseline SLMs hallucinated far less compared to the others.

\begin{figure}[H]
    \centering
    \includegraphics[width=\linewidth]{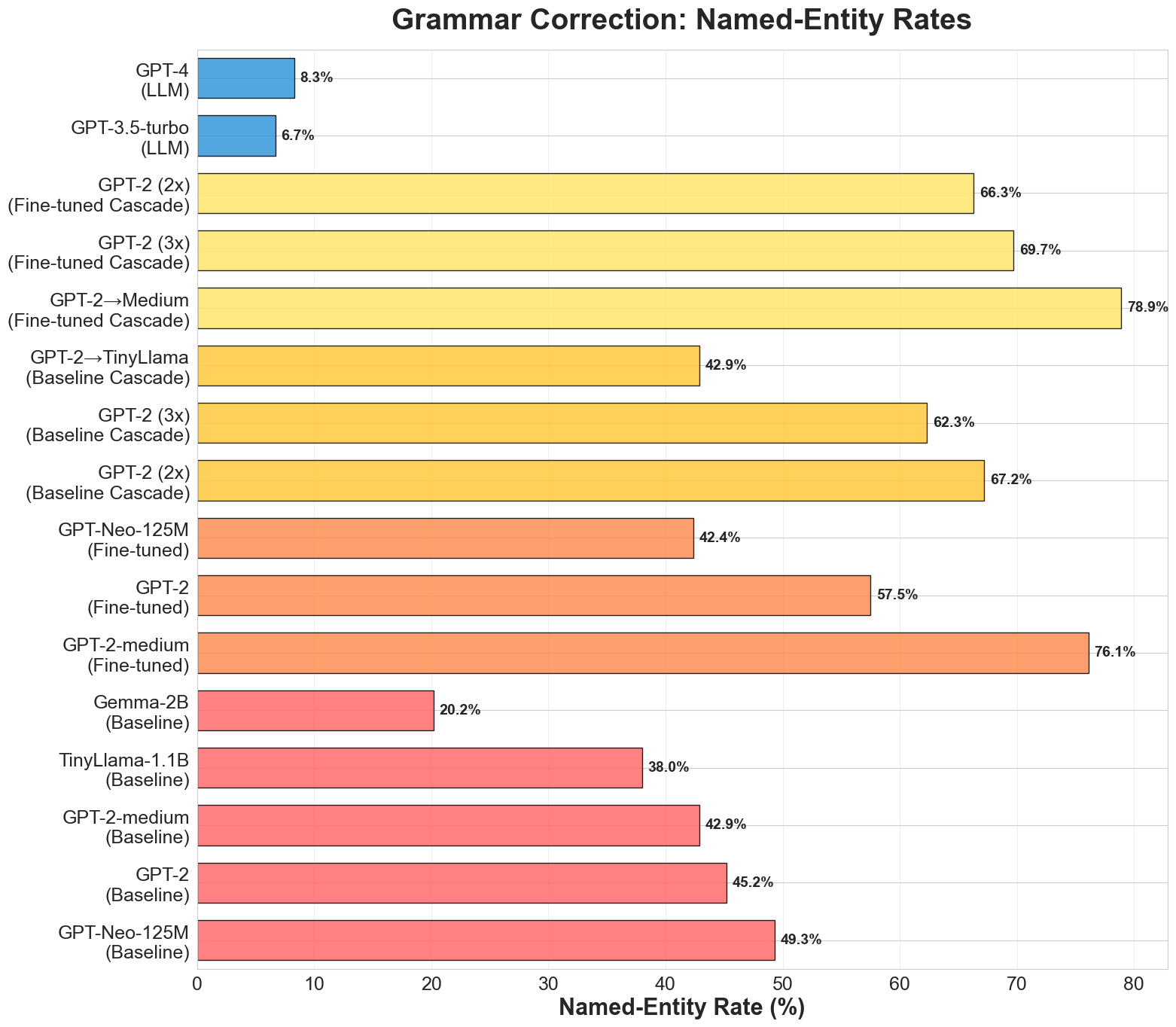}
    \caption{Grammar Correction: Named Entity Rates}
    \label{fig:mylabel}
\end{figure}

Figure 12 shows how much all of the implementations hallucinated when simplifying text. As you can see, just like in grammar correction, the LLMs and baseline SLMs hallucinated far less compared to the others.

\begin{figure}[H]
    \centering
    \includegraphics[width=\linewidth]{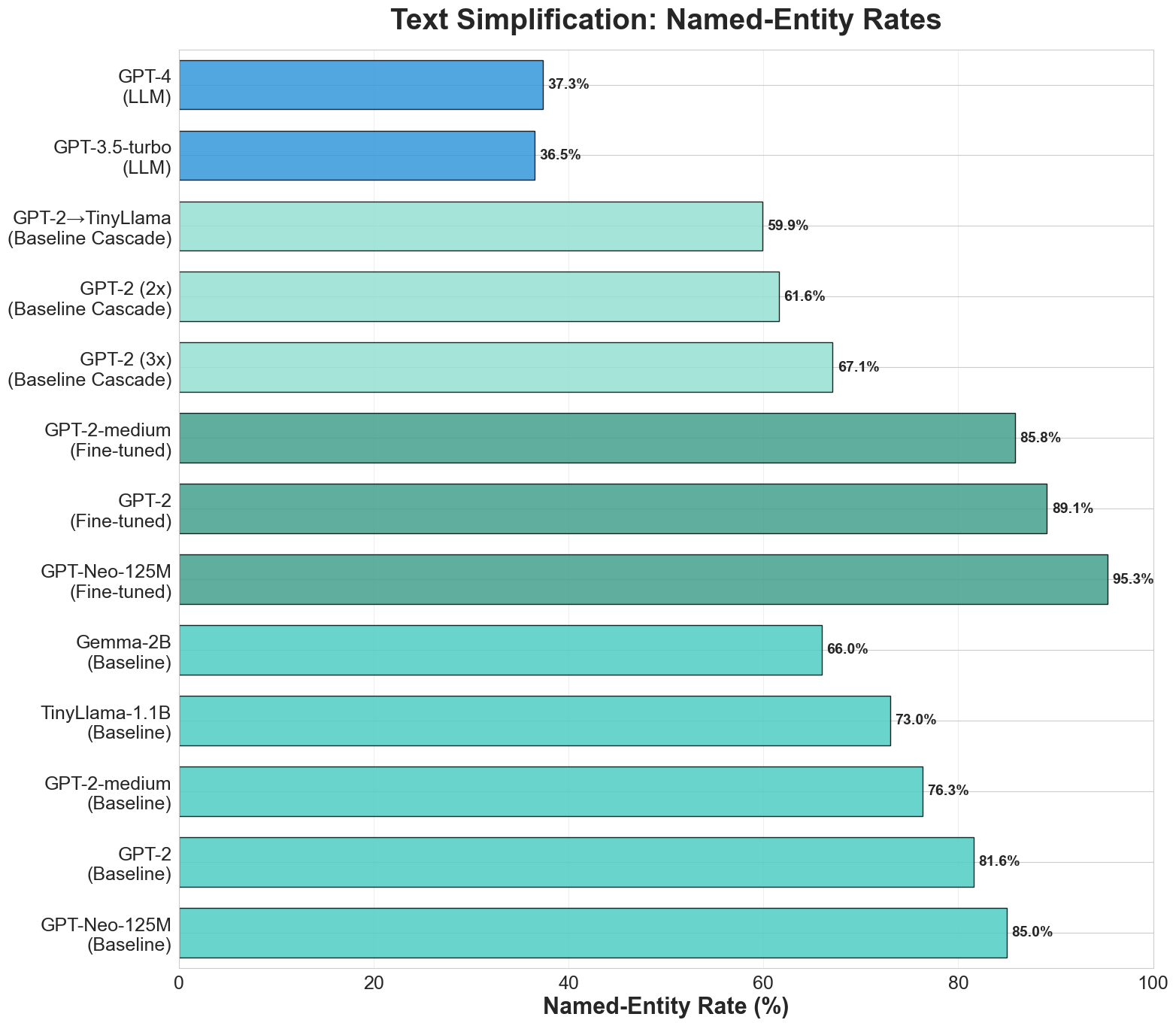}
    \caption{Text Simplification: Named Entity Rates}
    \label{fig:mylabel}
\end{figure}

\section{Discussion}

In the context of grammar correction, SLMs are far behind LLMs. This is seen as GPT-3.5 Turbo scored around 75 in GLEU with an $F_{0.5}$ score of around 51. The best performing SLM was Gemma 2B, who only achieved around a GLEU of 25 and an $F_{0.5}$ score of 16. This is around three times worse. On top of this, the average SLM $F_{0.5}$ score was only around 5, meaning that most SLMs failed the JFLEG set miserably. It is also important to note that the SLMs had much higher NERs on average than the LLMs. When comparing our work to prior evaluations on JFLEG, we observe that human performance and strong models such as Fluency Boost achieve scores in the low 60s, while many other neural and statistical baselines score in the 40s and 50s \cite{ge2018fluency}. Our SLMs (baseline, fine-tuned, and cascaded) performed significantly worse than this, indicating that small decoder-only models struggle with fine-grained grammar correction. We also noted that the LLMs outperformed the past research.

On the other hand, SLMs performed much better in the context of text simplification. When comparing SARI scores, Gemma 2B scored around 55, which was only about 4 points behind GPT-3.5 Turbo, which scored around 59. SLMs also performed comparably in terms of FRE where GPT-Neo 125M was the best overall performer. While these results may initially be seen as extremely impressive, as Gemma 2B is over 80 times smaller than GPT-3.5 Turbo, it is important to note that SLMs (including Gemma) performed much worse in terms of compressing text. Essentially, while their responses may have been simplified well, they were often much longer than the original input text. It is also important to mention that because of all this extra text, the SLMs were graded as hallucinating moderately more than the LLMs. When relating to previous evaluations on ASSET, we see encoder-decoder models such as MUSS and few-shot LLMs such as Flan-U12 achieve SARI scores in the low 40s. While some of our SLMs outperform them (and are much smaller), our fine-tuned models fall below them \cite{kew2023bless}. It is also important to note that the other models in prior research typically compress and maintain meaning better \cite{kew2023bless}.

When fine-tuning was attempted, it helped in some areas, but was overall harmful. In grammar correction, SLM $F_{0.5}$ scores were almost identical both before and after fine-tuning. In text simplification, the fine-tuned SLMs scored worse (SARI) than their baselines. This suggests overfitting and unnecessary training. Based on the results, fine-tuning felt very fragile and not very helpful. The fine-tuned models also hallucinated moderately more than the baselines in both grammar correction and text simplification according to our NER metric.

Overall, we found that model size made a large difference, especially in the SLMs, as Gemma 2B and TinyLlama 1.1B usually outperformed the smaller GPT models. However, even Gemma 2B's performance was well below that of the LLMs. Also, in terms of hallucination and the ability to make specific, fine-grained edits, the LLMs were far superior. On top of this, we found that the text simplification task as a whole was much more hallucination-prone than the grammar correction task, likely because in grammar correction, only small edits were ideal. While SLMs may technically be more efficient in terms of their cost and size, the quality is simply not good enough to make it worth it in most cases.

\section{Conclusion}

After exploring whether fine-tuned decoder-only SLMs could approach or replace LLMs in grammar correction and text simplification, we found that LLMs are currently not replaceable. While certain SLMs perform quite well for their size, their overall performance remains significantly below baselines on key metrics, such as SARI and GLEU, when considering compression and hallucinations. Even when fine-tuning or applying strategies such as cascading, the results were inconsistent and often worsened due to hallucinations and extra text. LLMs, on the other hand, performed substantially better, even exceeding many published results. However, SLMs are cheaper and perform more efficiently when taking their size into account, but the reality is that the quality gap limits real-world deployment. For future research, looking at mid-sized models may be ideal as they may perform much better and would still be much smaller than today's LLMs. Overall, our results suggest that while fine-tuned SLMs are appealing for efficiency, substantial modeling and training advances are still needed before they can replace large language models for high-quality rewriting.

\section{Group Member Contributions}

\subsection{Anthony Lamelas}
In the early stages of the project, Anthony worked on refining the idea and finding which datasets to train and test on. Anthony also attended and led the mentor meetings. He also organized meetings among group members and planned out tasks. When it came to the development of the scripts and the system, Anthony wrote the vast majority of the code and all of the documentation. On top of this, Anthony also wrote the majority of the first draft and final draft (contributed to every single section in both of the drafts). He wrote the entirety of the abstract, introduction, data, methodology, experiments, and group member contribution sections on both drafts. He also wrote all of the appendices. Lastly, Anthony contributed a fair portion to the project proposal and presented the project.

\subsection{Ed Ye}
Ed attended and participated in all of the group meetings, worked on evaluating some of the grammar models, and wrote some of the first draft and final report. Ed's main contributions on the development side were evaluating Gemma 2B and TinyLlama 1.1B. He also created the visualizations for them and wrote about them in the results, discussion, and conclusion sections. Ed also contributed a fair amount to the proposal.

\subsection{Sandra Cai} 
Sandra worked on related work and the proposal, and went to group meetings.

\subsection{Jeremy Luo}
Jeremy wrote some evaluation code, did research, and went to group meetings.

\subsection{Miki Osada}
Miki went to group meetings and contributed to the proposal.

\bibliography{custom}
\appendix
\section{Hyperparameters}
\label{sec:appendix}

\subsection{Training Hyperparameters}
The training hyperparameters were set as follows:

\begin{itemize}

    \item \textbf{Epochs:} 3
    \item \textbf{Optimizer:} AdamW
    \item \textbf{Learning Rate:} 5e-5
    \item \textbf{Warmup:} 500 steps (linear warmup)
    \item \textbf{Max Sequence Length:} 512 tokens
    \item \textbf{Gradient Accumulation:} 4 steps
    \item \textbf{Weight Decay:} 0.01

    \item \textbf{Grammar Models:}
    \begin{itemize}
        \item \textbf{Batch Size:} 8
        \item \textbf{Save Steps:} 500
    \end{itemize}

\item \textbf{Text Simplification Models}
\begin{itemize}
        \item \textbf{Batch Size:} 12
        \item \textbf{Save Steps:} 1000
    \end{itemize}

    \item \textbf{Efficiency Settings:} bfloat16 or float16 if unsupported
    \begin{itemize}
        \item Gradient checkpointing for GPT-2 Large \& GPT-Neo-1.3B
    \end{itemize}

    \item \textbf{Prompt Format:}
    \begin{itemize}
        \item Grammar: \texttt{"Correct this text": [input] \textbar\ Corrected: [target]}
        \item Simplification: \texttt{"Simplify this text": [input] \textbar\ Simplified: [target]}
    \end{itemize}

\end{itemize}

\subsection{Inference Hyperparameters}

After observing very high hallucination rates on the validation set at a temperature of 0.7, all final experiments used a greedy decoding strategy. The maximum new tokens was also restricted to 60 and 80 for grammar and text simplification, respectively. The batch size was set to 8 for evaluation throughput.

\section{Hardware and HPC Setup}
\label{sec:appendix}
The fine-tuning jobs were allocated 2 GPUs and 8 CPU cores per task. Grammar jobs were allocated 32 GB of RAM, and text simplification jobs were allocated 64 GB. The evaluation jobs were allocated 1 GPU, 4 CPU cores, and 16 GB of RAM. All jobs had a time limit of 72 hours.

Python scripts were used for training and evaluation. SLURM scripts were also used to submit and run the Python scripts as jobs on the HPC.

\section{Training and Implementation Challenges}
\label{sec:appendix}
A large amount of time was spent dealing with file not found, environment not found, and module not found errors. This was largely due to our group's lack of experience with the HPC. These errors cost us a large amount of time as some jobs would run for a few hours, then fail. We also had some delays due to not giving jobs enough batches, time, or memory, causing us to re-run many jobs. The reason why these delays were so costly is that feedback when submitting a script is not instant and often takes multiple hours. 

\section{Fine-Tuned Model Validation Data}
\label{sec:appendix}
Figure 13 shows the NER for the models that we fine-tuned. As you can see, the models were hallucinating often, partially due to the use of a temperature of 0.7.

\begin{table}[h]
\centering
\small
\begin{tabular}{l r}
\hline
\textbf{Model} & \textbf{NER} \\
\hline
GPT-2 (Gram.)            & 70.1 \\
GPT-2 Medium (Gram.)     & 83.8 \\
GPT-Neo-125M (Gram.)     & 56.7 \\
GPT-2 (Text Simp.)           & 83.2 \\
GPT-Neo-125M (Text Simp.)    & 95.4 \\
\hline
\end{tabular}
\caption{Validation-set Hallucination Rates (Gram. = Grammar, Text Simp. = Text Simplification)}
\label{tab:hpc_hallucination}
\end{table}

\end{document}